**Knowledge Technology**

# Fact Sheet

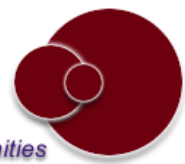

# Semantic Web

## What is the Semantic Web?

"The Semantic Web is an extension of the current web in which information is given well-defined meaning, better enabling computers and people to work in cooperation"[1]. The key enabler of the Semantic Web is the need of many communities to put machine-understandable data on the Web which can be shared and processed by automated tools as well as by people. Machines should not just be able to display data, but rather be able to use it for automation, integration and reuse across various applications.

The European Commission is funding numerous projects related to Ontologies and the Semantic Web; even more will be funded in its recently launched Sixth Framework Research Programme. The worldwide Semantic Web community is growing rather fast and forces are being joined with other technology developments such as Web Services or multimedia. Last, but not least, vendors are already offering mature products and solutions based on semantic technologies. Thus, the Semantic Web is currently moving from being a vision to becoming reality.

## How could the Semantic Web look like?

Even worse: "How would you explain the Semantic Web to your grandparents?" Answering this question is the challenge for participants of the Semantic Web Challenge[2]. Out of 10 participants the winners of the Challenge 2003 have been presented at the 2[nd] International Semantic Web Conference[3]. It might be questionable whether grandparents would understand the aim of the winning application, viz. that it "combines information from multiple heterogeneous sources, such as published RDF sources, personal web pages, and data bases in order to provide an integrated view of this multidimensional space"[4]. Nevertheless, it offers the flavour of current Semantic Web technologies.

A very illustrative and at the same time amusing article gives a glimpse into the far future, viz. "August 2009: How Google beat Amazon and Ebay to the Semantic Web"[5].

## Why is the Semantic Web important?

To illustrate the potential importance of the Semantic Web we will start with some quotes from "important people and institutions".

- "The way software and devices communicate today is doomed. To interoperate on the X Internet, they'll use decentralized data dictionaries based on emerging Semantic Web technologies."

  *David Truog, Forrester Report: "How the X Internet Will Communicate", December 2001.*

- "While the industry is busy creating the underpinnings of open computing with standards like eXtensible Markup Language, still **missing are what Plattner calls 'semantic' standards**, or how to make different computers recognize data about a business partner, a customer, or an order and know what to do with it. In other words, said Plattner, the software industry is **building an alphabet but hasn't yet invented a common language**."

  *Hasso Plattner, SAP, in CNet News, 27[th] of March, 2002.*

We will elaborate more on the mentioned issues in the following subsections.

---

1 Berners-Lee, T., Hendler, J., & Lassila, O. (2001). The semantic web. Scientific American, 2001(5). available at http://www.sciam.com/2001/0501issue/0501berners-lee.html
2 Semantic Web Challenge, initiated in cooperation with the International Semantic Web Conference 2003, to be continued in 2004, cf. http://challenge.semanticweb.org/
3 International Semantic Web Conference, cf. http://iswc.semanticweb.org/
4 CS AKTiveSpace Tour, cf. http://triplestore.aktors.org/SemanticWebChallenge/
5 P. Ford. August 2009: How Google beat Amazon and Ebay to the Semantic Web. Published Friday, July 26, 2002. Online available at http://www.ftrain.com/google_takes_all.html

# About the Semantic Web

**History**. The advent of the World Wide Web (WWW) gave mankind an enormous pool of available information. The WWW is based on a set of established standards, which guarantee interoperability at various levels: e.g., the TCP/IP protocol provides a basis for transportation of bits, on top HTTP and HTML provide a standard way of retrieving and presenting hyperlinked text documents. Applications could easily make use of this basic infrastructure, which led to the now existing WWW. However, nowadays the sheer mass of available documents and the insufficient representation of knowledge contained in documents makes "finding the right things" real work for human beings. A major shortcoming of HTML is that it is well suited for human consumption, but not for machine-processability. As such, to interpret the information given in documents the human has always to be in the loop.

To overcome such shortcomings, ontologies recently have become a topic of interest in computer science. Ontologies provide a shared understanding of a domain of interest to support communication among human and computer agents, typically being represented in a machine-processable representation language. Thus, ontologies are seen as key enablers for the Semantic Web. Refer to the accompanying Knowledge Technology Fact Sheet on Ontologies (cf. http://www.ktweb.org/doc/Factsheet-Ontologies-0306.pdf) for more information.

Ontobroker (initially developed at the Institute AIFB/University of Karlsruhe, and now commercialized by the company Ontoprise) and SHOE (University of Maryland) were two ontology-based systems ahead of their time. Both systems relied on additional semantic markup which was put into regular web pages, so-called annotations. The systems showed very early the feasibility of adding machine-processable semantics to web pages. Many ideas of this work made it into current Semantic Web standards of the W3C (see the later section on Standards).

Both systems also heavily influenced the development trends of semantic technologies. In the following we will briefly characterize typical Semantic Web tools and give examples of existing commercial and academic tools. It is quite noteworthy that most tools are currently not only being used to build and maintain WWW applications, but also corporate intranet solutions.

**Ontology Editors** allow for creation and maintenance of ontologies, typically in a graphically oriented manner. There exists a plethora of available implementations, each having its own specialty and different functionalities. Common to most editors is the ability to create a hierarchy of concepts (such as "Car is a subconcept of Motor Vehicle") and to model relationships between those concepts (such as "A car is driven by a person". More advanced editors also allow the modelling of rules, but to explain this is beyond the scope of this paper.

*OntoEdit* is the most prominent commercial ontology editor (available from http://www.ontoprise.com). Unlike most other editors, OntoEdit comes with a strong inferencing backbone, Ontobroker, which allows the modelling and use of powerful rules for applications. Numerous extensions, so-called plugins, exist to adapt OntoEdit to different usage scenarios such as database mapping. Last, but not least, a full-fledged tool support is provided by the Ontoprise team which makes it attractive for companies.

*Protégé* is the most well-known academic ontology editor with a long development history (available from http://protege.stanford.edu/). Similar to OntoEdit it is based on a flexible plugin framework. Numerous plugins have been provided so far which nicely demonstrate possible extensions for typical ontology editors. An example is the PROMPT plugin, which allows for merging of two given ontologies into a single one.

*KAON* (http://kaon.semanticweb.org) is not only an ontology editor, but rather an open-source ontology management infrastructure targeted at business applications. It includes a comprehensive tool suite allowing easy ontology creation and management, as well as building ontology-based applications. An important focus of KAON is on integrating traditional technologies for ontology management and application with those used in business applications, such as relational databases.

**Annotation Tools** allow for adding semantic markup to documents or, more generally, to resources. The great challenge here to automate the annotation task as much as possible to reduce the burden of manual annotation for large scale resources. A good place to find further information on annotation and authoring, a quite related topic, is http://annotation.semanticweb.org/.

OntoMat-*Annotizer* (available from http://annotation.semanticweb.org/ontomat) is currently the most prominent annotation tool. It is based on a full-fledged annotation framework called CREAM, which is already being extended to support semi-automatic annotations of documents as well as annotation of databases.

**Inference Engines** allow for the processing of knowledge available in the Semantic Web. In a nutshell, inference engines deduce new knowledge from already specified knowledge. Two different approaches are applicable here: having general logic based inference engines, and specialized algorithms (Problem Solving Methods). Using the first approach one can distinguish between different kinds of representation languages such as Higher Order Logic, Full First Order Logic, Description Logic, Datalog and Logic Programming (cf. http://semanticweb.org/inference.html). Recently, in a contest-like project three state-of-the-art inference engines were evaluated with quite interesting results (cf. http://www.projecthalo.com/). Inference engines are *per se* very flexible and adaptable to different usage scenarios such as information integration or, to show the bandwidth of possible scenarios, intelligent advisors.

*Ontobroker* (http://www.ontoprise.com) is the most prominent and capable commercial inference engine. It is based on Frame Logic, tightly integrated with the ontology engineering environment OntoEdit and provides connectors to typical databases. It was already used in numerous industrial and academic projects.

*FaCT* is one of the most prominent Description Logics based inference engines (available from http://www.cs.man.ac.uk/~horrocks/FaCT/). In a nutshell, FaCT (Fast Classification of Terminologies) is a Description Logic classifier that can also be used for modal logic satisfiability testing.

## Research activity

The OntoWeb thematic network has currently over 100 partners coming from academia and industry. Most of them are located in Europe, but there also exist strong links to communities in the United States of America and Asia. The goal of the OntoWeb thematic network is to bring researcher and industrials together, enabling the full power ontologies may have to improve information exchange in areas such as: information retrieval, knowledge management, electronic commerce, and bio-informatics. The corresponding OntoWeb.org portal (http://www.ontoweb.org) (i) provides numerous documents (deliverables) with state-of-the-art overviews on ontology technologies and (ii) is itself a Semantic Portal relying on ontology technology[6].

Currently ongoing European projects include "SWAP – Semantic Web and Peer to Peer" (http://swap.semanticweb.org) and "WonderWeb – Infrastructure for the Semantic Web" (http://wonderweb.man.ac.uk). Upcoming, but currently still being negotiated, are several large and medium European funded projects, e.g. "SEKT – Semantically Enabled Knowledge Technologies" (http://sekt.semanticweb.org).

The most prominent conference is the annual "ISWC –International Conference on Semantic Web (http://iswc.semanticweb.org), but there are numerous Semantic Web related tracks and workshops at major conferences such as the IJCAI (http://ijcai.org/) or the WWW (http://www2003.org/). A more complete project and technology list and further information about the ideas of the Semantic Web can be found at the Semantic Web.org portal (http://www.semanticweb.org).

An interesting number to mention: instead of envisioned 250 people, over 400 people attended the ISWC 2003. Thus, the number of attendees, both researchers and industrials, almost doubled in comparison to 2002!

## Standards activity

Standards activities for Semantic Web languages are mainly driven by working groups of the W3C (http://www.w3c.org/). The Semantic Web layer cake by Tim Berners-Lee shows the layering of the current state-of-the-art and future planned standards; on the right side can be seen the current status of each layer. While XML as a baseline allows for a syntactical description of documents, the layers RDF, Ontology and Logic are adding machine-processable semantics - a necessary prerequisite for, for example, sharable web resources:

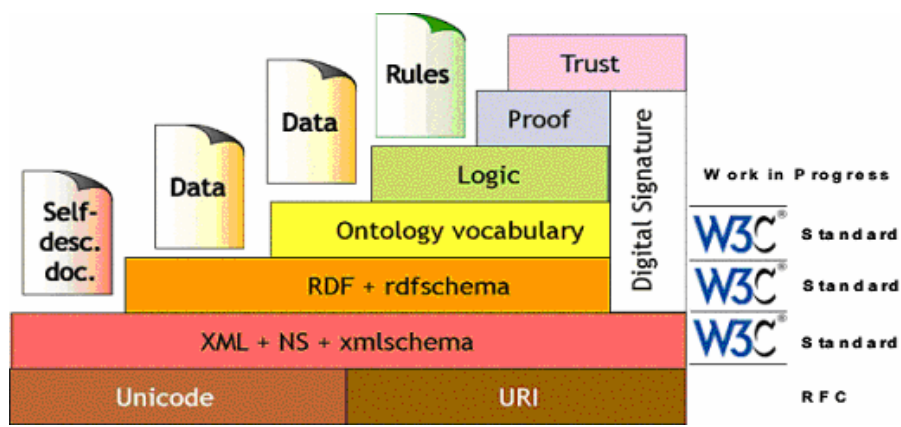

---

[6] Spyns, P., Oberle, D., Volz, R., Zheng, J., Jarrar, M., Sure, Y., Studer, R., & Meersman, R. (2002). OntoWeb - a semantic web community portal. In: Karagiannis, D. & Reimer, U. (Eds.) (2002). Proceedings of the Fourth International Conference on Practical Aspects of Knowledge Management (PAKM2002), volume 2569 of Lecture Notes in Artificial Intelligence (LNAI), Vienna, Austria. Springer, pages 189-200.

On top of the core standards for XML (eXtensible Markup Language) and RDF (Resource Description Framework) the W3C WebOnt working group (http://www.w3.org/2001/sw/WebOnt) has recently released the OWL web ontology language standard (http://www.w3.org/TR/owl-ref). Future work remains to be done for the Logic, Proof and Trust layers.

## Commercial activity

Ontoprise GmbH (http://www.ontoprise.com) is, according to Gartner, the leading technology provider for Semantic Technologies. They offer a broad spectrum of semantic technologies, e.g. (i) OntoEdit, a flexible and modularly extensible ontology engineering environment, (ii) Ontobroker, a full-fledged inference engine to process multiple ontologies, typically used as a semantic middleware server to provide integrated access to various data sources, and (iii) SemanticMiner, a knowledge retrieval search and browsing application on top of Ontobroker which combines information retrieval with ontology technology. Further companies on the market are, for example, Network Inference, Ontology Works and Intellidimension.

## Current issues

Some basic research issues still have to be solved especially for the creation and handling of ontologies such as e.g. the evolution of ontologies and metadata or complex mappings of multiple ontologies, to name but a few. To support research in this fast growing community, the European Commission's Sixth Framework Programme (http://fp6.cordis.lu/fp6/) included *"Semantic-based knowledge systems"* as a strategic research objective in the IST (Information Society Technologies) area, with proposed funding of 55 MEuro.

Baseline Semantic Web Standards like XML, RDF and OWL are published. The Semantic Web technologies are moving from being scientific prototypes to being full-fledged commercial products. Ontologies become available as shared, re-usable entities for specific domains.

On top, Semantic Web technologies are in the transition of merging with other technologies such as Web Services (towards "Semantic Web Services") and the GRID (towards a "Semantic GRID"). Not to forget the Google people, but … who knows what they are doing?

**The Semantic Web is moving from being a vision to becoming reality.**